\documentclass[crcready]{iosart2x}

\usepackage{amsmath}
\usepackage[group-separator={,}]{siunitx}
\usepackage{booktabs}
\usepackage{graphicx}

\pubyear{0000}
\volume{0}
\firstpage{1}
\lastpage{1}

\begin{document}

\begin{frontmatter}

\title{How Stable is Knowledge Base Knowledge?}
\runtitle{KB Stability}

\author[A]{\inits{}\fnms{Suhas} \snm{Shrinivasan}\ead[label=e1]{}},
\author[B]{\inits{}\fnms{Simon} \snm{Razniewski}\ead[label=e2]{srazniewski@mpi-inf.mpg.de}}
\address[A]{\orgname{University of Göttingen}}
\address[B]{\orgname{Max Planck Institute for Informatics}}

\begin{abstract}
\textsc{Knowledge Bases} (KBs) provide highly structured semantic representation of the real-world in the form of extensive collections of facts about real-world entities, their properties and relationships between them. They are ubiquitous in large-scale intelligent systems that exploit structured information such as in tasks like structured search, question answering and reasoning, and hence their data quality becomes paramount. The inevitability of \textit{change} in the real-world, brings us to a central property of KBs --- they are highly dynamic in that the information they contain are constantly subject to change. In other words, KBs are \textit{unstable}. 

In this paper, we investigate the notion of \textit{KB stability}, specifically, problem of KBs changing due to real-world change.
Some entity-property-pairs do not undergo change in reality anymore (e.g., \textit{Einstein-children} or \textit{Tesla-founders}), while others might well change in the future (e.g., \textit{Tesla-board member} or \textit{Ronaldo-occupation} as of 2021). This notion of real-world grounded change is different from other changes that affect the data only, notably correction and delayed insertion, which have received attention in data cleaning, vandalism detection, and completeness estimation already. 
To analyze KB stability, we proceed in three steps. (1) We present heuristics to delineate changes due to world evolution from delayed completions and corrections, and use these to study the real-world evolution behaviour of diverse Wikidata domains, finding a high skew in terms of properties. (2) We evaluate heuristics to identify entities and properties likely to not change due to real-world change, and filter inherently stable entities and properties. (3) We evaluate the possibility of predicting stability post-hoc, specifically predicting change in a property of an entity, finding that this is possible with up to 83\% F1 score, on a balanced binary stability prediction task.

\end{abstract}


\end{frontmatter}




\section{Introduction}


Knowledge bases (KBs) like Wikidata, DBpedia~\cite{dbpedia} or YAGO~\cite{YAGO} are cornerstones of the Semantic Web~\cite{hogan2020knowledge}, and increasingly used as backbones for AI applications like question answering and dialogue~\cite{dubey2019lc,abujabal2017quint,industry-kg}. A crucial task for KB curators as well as consumers is to understand their data quality, as quality issues propagate throughout use cases and adversely affect user experience~\cite{zaveri2016quality, paulheim2017knowledge}. 

The goal of this paper is to understand the dimension of \textbf{KB stability}: Can we trust data to remain as is, that was last curated at a certain point in time in the past?

\noindent\textbf{Motivation.\ \ }
Stability has high relevance in web-scale KBs: For example, Wikidata entities are often created in a concerted, one-off effort, then left untouched for months or years. And even if KB content is updated regularly, applications may rely on discrete dumps that risk going stale. The QAnswer system \cite{diefenbach2019qanswer}\footnote{\url{https://qanswer-frontend.univ-st-etienne.fr}}, for instance, still responds to the question ``Who is the president of the USA'' with ``Donald Trump'' (as of Mar 11, 2021). Knowledge about stability is also crucial for negative knowledge \cite{arnaout2021negative}, as real-world evolution may render it incorrect.

Knowledge about KB (in-)stability could help both consumers and curators. KB consumers, on the one hand, could use this information to qualify the reliability of their applications, e.g., annotating the query answer for Musk's residence with a warning like, ``\textit{This information is volatile and may be outdated!}''. KB curators, on the other hand, could use this information to efficiently allocate their update-efforts - among a trove of entity-property-pairs, identifying those most likely to have undergone change since the last review.


\noindent\textbf{State of the art and its limitations.\ \ }
Several works have looked at estimating KB data \textit{quality}. Identifying inaccuracies in the data~\cite{dong2014knowledge, lehmann2010ore, ma2014learning, li2015probabilistic}, possibly detecting vandalism~\cite{wikidata-vandalism}, is one such direction. Another direction is identifying (in-)completeness in KBs~\cite{galarraga2016completeness, galarraga2013amie, factorize-YAGO, luggen-completeness, balaraman2018recoin}. Both correction and completion also change KB content, yet, in contrast to real-world evolution, are intentionally performed, and with the goal of improving KB quality. Real-world evolution, in contrast, is a process outside the control of KB curators, by which utility of a KB snapshot continuously erodes.

Currently, KBs like Wikidata, DBpedia and YAGO can provide temporal qualifiers along with assertions (e.g., \textit{Ronaldo, team, Juventus, start:2019}), which provide a coarse hint as to where real-world change is possible. 
A few works have tackled related issues, such as efficiently distributing update resources to maximize KB freshness w.r.t.\ a source website~\cite{chinese-kb}, or analyzing the possibility to link KB changes to verb changes in encyclopedic texts~\cite{wijaya2015spousal}. Similarly, work in~\cite{hao2020outdated} distributes bulk updates based on coarse estimates of change likelihoods, and change interrelations that allow to formulate a set-cover style problem.
Other works have looked at optimizing web re-crawl resource distributions based on estimations of the frequency of change of web sources~\cite{change-cho-molina}, or under cost-benefit tradeoffs~\cite{razniewski2016optimizing}. None of these works specifically target the problem of predicting KB stability.

\noindent\textbf{Approach and contribution.\ \ }
Our guiding research question is: \textit{``How can we identify whether KB content is stable, w.r.t.\ real-world changes?''} Specifically, we aim to qualify \textit{entity-property-pairs} by their likelihood to have undergone real-world change since its latest trusted edition. For example, if \textit{Tesla}'s \textit{board members} property were last reviewed in 2010, it is likely that they underwent real-world change since then, yet that is not the case for its property \textit{founders}. Similarly, what \textit{team} \textit{Ronaldo} is playing for may well undergo real-world change within short time, but this less likely for properties like \textit{citizenship} or \textit{first name}.

We break the guiding research question down further into three sub-questions:
\begin{enumerate}
    \item [\textbf{RQ 1.}] How frequent is KB change due to real-world change, and how can we differentiate it from other types of change?
    \item [\textbf{RQ 2.}] How can we identify subject-property-pairs that are stable?
    \item [\textbf{RQ 3.}] For subject-property-pairs that are unstable, how can we predict the change likelihood?
\end{enumerate}
While tackling RQ 1, we encounter the challenge of automatically distinguishing KB changes resulting from real-world changes from changes resulting from delayed completion. Delayed completions, e.g., adding in 2021 information about \textit{Einstein}'s \textit{education}, and corrections, e.g. changing \textit{Einstein}'s \textit{country of birth} from \textit{USA} to \textit{Germany}, generally dominate the number of changes in a KB. To distinguish between changes due to real-world change and delayed completions, we explore three heuristic criteria --- (i) using the partial completeness assumption~\cite{galarraga2013amie} and (ii) bulk updates to identify delayed completions, and (iii)  using time-stamp information. Although each of the heuristics has limitations, we show that in practice they are effective, jointly giving good insights into the real-world change behaviour underlying KBs.

In RQ 2, we investigate whether there are effective filters that allow to rule out certainly stable information. For example, dead entities should hardly undergo change, and properties such as \textit{first name} or \textit{place of birth} should also be stable. We investigate heuristics for identifying stable entities (e.g., dead ones or Wikipedia-inactive ones), and for stable properties (e.g., single-valuedness or having had less historic edits). Our results show that with regard to entities, generalizations of death are indeed a useful feature, while stability in Wikipedia has almost no relation to real-world stability, and with regard to properties, single-valuedness indicates stability whereas number of historic edits is useless as far as real-world changes are concerned.

In RQ 3, we approach stability estimation as a (property) prediction problem. We analyze which features hold most predictive power, finding in particular that textual features from encyclopedic source, and their change thereof, can predict stability with up to 83\% F1 score considering a 3-year window, providing interesting interpretable insights into entity characteristics implying change.

Our novel technical contributions include:
\begin{enumerate}
    \item Introduce and formalize the notion and problem of KB stability;
    \item Introduce and evaluate heuristics for identifying KB changes resulting from real-world changes, among general KB changes;
    \item Introduce and evaluate heuristics to determine stable properties and stable entities;
    \item Analyse and evaluate features for supervised prediction of property stability.
\end{enumerate}

\renewcommand{\paragraph}[1]{\vspace{1.5mm}
\noindent
{\bf #1.\ }}

\paragraph{Delineation} Stability in general, like other typical data quality dimensions~\cite{pipino2002quality} such as timeliness, completeness and correctness, is desirable. But, unlike the other dimensions, stability cannot be controlled for by the KB curator, in that a KB cannot be ``made" stable - it results from the whims of the reality. KBs can be timely, complete and correct at one point in time, but due to the changing world, the data in the KB at a later point in time is bound to become stale due to the changing world, hence motivating the problem of stability prediction.

\section{Related Work}


\sloppy
\paragraph{KB Completeness} The basis for incompleteness in KBs is the Open World Assumption (OWA), which states that any assertion that the KB does not contain, is not necessarily \textit{false}, but simply \textit{unknown}. Most KBs assume OWA, and is naturally the most realistic assumption, but also makes automatically determining completeness hard. Work in~\cite{galarraga2017predicting} systematically analyzes completeness in KBs and discusses relaxed assumptions, a.k.a. \textit{completeness oracles}, that guess whether a given \textit{entity-property} pair in a KB is complete, such as partial completeness assumption (PCA), which states that if an \textit{entity-property} pair has at least one value, e.g., $\textit{partnerOf}\left(\textit{Don},\textit{Melania}\right)$, then the \textit{entity-property} pair (\textit{Don-partnerOf} or \textit{Melania-partnerOf}) is complete. More sophisticated oracles are also proposed to predict incompleteness and even perform completion based on association rule mining such as AMIE~\cite{galarraga2013amie}, that extracts \textit{Horn} clauses based on their support in the KB, e.g., $\textit{sonOf}\left(\textit{Don},\textit{Barron}\right)\land\textit{sonOf}\left(\textit{Melania},\textit{Barron}\right)\implies\textit{spouseOf}\left(\textit{Don},\textit{Melania}\right)$. Completions are also approached using latent feature models, a.k.a. embedding models, which assume assertions of a KB to be conditionally independent of each other given a set of global latent features, and use matrix factorization or analogous techniques on the KBs in order to learn these latent features, which are eventually used to score any arbitrary \textit{subject-property-object} triple signifying its likelihood of being true~\cite{nickel2012factorizing, bordes2013translating, socher2013reasoning}. Another work looks at class completeness, e.g., "Does the KB have a complete list of volcanoes" and uses non-parametric estimation techniques, such as the Jackknife estimator, inspired from species-sampling~\cite{luggen-completeness}. Relative completeness of an entity is another idea approached using comparison with data present for similar entities~\cite{balaraman2018recoin}. There also exist work that estimates completeness based on coverage in text, e.g., if there exists the sentence, "Don was married twice.", then the scalar implication is that entity \textit{Don} should have two values for \textit{Don-spouse} pair~\cite{razniewski2016optimizing}.

\paragraph{Temporal Completeness} Addressing static KB completeness allowed us to ask completeness questions w.r.t \textit{entity-property} (to check if all subjects/ objects specific to the pair are contained in the KB) or w.r.t \textit{entity-entity} pairs (to check if all properties or relations between the two entities are contained in the KB). Addressing tKB completeness allows us, theoretically, to ask completeness questions with an additional dimension of timestamp, such as if the KB has a complete list of objects for an \textit{entity-property} pair \textit{in 2015}. Several works extend static KB completeness to tKBs, such as extending embedding models to account for time information - by using integer linear programming with temporal consistency as constraints~\cite{jiang2016towards}, by incorporating temporal representations directly into the scoring functions of static embedding models~\cite{leblay2018deriving, ma2019embedding}. Recent works have considering modeling temporally evolving KBs using sophisticated temporal point processes~\cite{trivedi2017know} and proposed for instance, the Neural Graph Hawkes process~\cite{han2020graph} that allows one to even consider future timestamps.

\paragraph{KB Accuracy} Another important dimension, apart from completeness, is correctness or accuracy. Works that are concerned with KB accuracy aim to develop methods that detect errors in KBs~\cite{paulheim2017knowledge}. Errors could manifest in entities or in properties. Work in~\cite{ma2014learning} target erroneous entities by using association rule mining to learn disjointness axioms that detect inconsistencies in entity type assertions. Work in~\cite{paulheim2014improving} exploits statistical distributions of entities and properties and flag errors when large deviations are detected. In fact~\cite{paulheim2014improving} targets both completeness and correctness together. Similarly, the embedding models discussed w.r.t completeness can also potentially used for corrections since given a triple, they essentially provide a score signifying likelihood for the existence of the triple~\cite{nickel2015review}. There also exist external methods such as \textit{DeFacto}~\cite{gerber2015defacto} that convert triples into natural language statements and look for its popularity in web searches. Furthermore, there exist work that check for accuracy in not just static KBs but also evolving KBs, e.g., work in~\cite{rashid2019quality} propose a quality assessment methodology for evolving KBs using proposed quality dimensions that are based on changes in KBs over subsequent releases, based on \textit{evolutionary analysis.}

\paragraph{Timeliness and stability} A few works have implicitly approached instability resulting from real-world change. For instance, work in~\cite{chinese-kb} attempts to synchronize KB with its growing encyclopedic source by learning an update frequency predictor of entities by using various features of the entity's encyclopedic page such as update frequency, popularity and centrality. Work in~\cite{wijaya2015spousal} similarly, learns \textit{verb} changes in natural language text in the encyclopedic source (Wikipedia article) that predict changes in corresponding structured information (Wikipedia infoboxes). Work in~\cite{hao2020outdated} tackles outdated fact detection in KBs by developing a human-in-the-loop framework and training a supervised classifier to perform automatic fact detection that uses as features a proxy for completeness of the subject entity of the fact, historical update frequency, number of links to the subject entity and so on, but unlike our work, it does not provide a deep enough analysis of the problem and trains a general model for all types of facts regardless of the type of property or the entity. Work that comes closest to ours is in~\cite{dikeoulias2019epitaph}, where stability of properties is predicted by training a supervised classifier that uses as features other properties of the entity. Other works have looked at optimizing web re-crawling based on estimation of frequency of change of web sources~\cite{change-cho-molina}, and also hypothesizing a cost-benefit framework~\cite{razniewski2016optimizing}. Yet another angle is provided by work that aims to predict the stability of natural language assertions via temporal tagging (intervals) \cite{kuzey2016time}, or via a coarse-grained classification into stability categories like \textit{few hours, few days, few weeks, ...}, based on linguistic features \cite{almquist2019towards}.

\section{Formalization} \label{sec:formalization}


\paragraph{Data model} We define a KB to be a set of facts that are quintuples. Assuming existence of infinite sets $E$ (\textit{entities}), $\Pi$ (\textit{properties}), $\Lambda$ (\textit{literals}), and $\Theta$ (\textit{timestamps}), we define a quintuple fact as $(s, p, o, t_v, t_a) \in {E}\times{\Pi}\times({E}\cup{\Lambda})\times{\Theta}\times{\Theta}$ where $s$ is the subject, $p$ the property, $o$ the object, $t_v$ timestamp representing valid time, i.e., the point-in-time at which the fact has occurred in the real-world, and $t_a$ timestamp is the transaction time, i.e., point in time the fact was added to the KB~\cite{tansel1993temporal}. $t_v$ can also be null. 

\paragraph{Subject-property-centric view} We conceptualize changes in KBs around properties of subject entities, e.g., \textit{Ronaldo} moves from \textit{Madrid} to \textit{Turin} in $2018$, and the KB is added with the fact $\left(\textit{Ronaldo, residence, \textit{Turin}, $2018$, $2018$}\right)$, and we say property \textit{residence} of subject entity \textit{Ronaldo} has changed. 




\sloppy \paragraph{Notation} In general, given two sampling timepoints $\tau_1$ and $\tau_2$, where $\tau_2 > \tau_1$, corresponding KB snapshots $\mathcal{K}_{\tau_1}$ and $\mathcal{K}_{\tau_2}$ and some \textit{subject-property} pair $(s, p)$, we denote the set of objects and corresponding valid timestamps and transaction timestamps belonging to $\left(s, p\right)$ in $\mathcal{K}_{\tau_1}$ as $O_{\mathcal{K}_{\tau_1}} = \{\left(o_i, t_v\left({o_i}\right), t_a\left({o_i}\right)\right)\}_{i=1}^{n}$, where $n \in \mathbb{N} \cup\{0\}$, and similarly for $\left(s,p\right)$ in $\mathcal{K}_{\tau_2}$, $O_{\mathcal{K}_{\tau_2}} =\{\left(o_i, t_v\left({o_i}\right), t_a\left({o_i}\right)\right)\}_{i=1}^{m}$, where $m \in \mathbb{N} \cup\{0\}$. For example, considering sampling timepoints $\tau_1 = 2017$ and $\tau_2 = 2020$; KB snapshots $\mathcal{K}_{2017}$ and $\mathcal{K}_{2020}$; object-timestamps sets $O_{\mathcal{K}_{2017}} = \{\left(Madrid, 2010, 2012\right)\}$ and $O_{\mathcal{K}_{2020}} = \{\left(Madrid, 2010, 2012\right), \left(Turin, 2018, 2018\right)\}$.

We say that a pair (s,p) is stable in an interval $[\tau_1,\tau_2]$, iff $O_{\mathcal{K}_{\tau_1}}=O_{\mathcal{K}_{\tau_2}}$.

\paragraph{Problem} Given a \textit{subject-property} pair $\left(s,p\right)$ and sampling timepoints $\tau_1$ and $\tau_2$, $\tau_2 > \tau_1$, what is the likelihood that \textit{p} is not stable between $\tau_1$ and $\tau_2$? In other words, what is the likelihood that \textit{p}, considering the time-interval, is stable?



\section{Change Analysis} \label{sec:change-analysis}


KBs undergo a lot of changes. For example, just in February 2021, 20 million edits were performed in Wikidata\footnote{\url{https://stats.wikimedia.org/\#/wikidata.org/contributing/edits/normal|bar|2019-11-25~2021-03-06|~total|monthly}}. But how many of these edits are due to real-world change? This number subsumes two other reasons for change - (i) delayed completions and (ii) corrections. In fact, any observable property change subsumes changes due to real-world change as well as delayed completions and corrections. Since we are interested in KB changes due to real-world change, it is important for us to develop methods that automatically identify if an observed KB change is due to real-world change, and consequently estimate for their prevalence among all changes that KBs undergo. But before we do so, we define a na\"{i}ve criterion for change which just measures some observable change in property values.

\sloppy \paragraph{Observable property change} Valid time is often not asserted in KBs, thus we next define observable property change, which is therefore more than just unstable KB content. We define \textit{observable property change} for property $p$ of subject entity $s$ between sampling timepoints $\tau_1$ and $\tau_2$ as the case when: $O_{\mathcal{K}_{\tau_1}} \neq O_{\mathcal{K}_{\tau_2}}$, i.e., there must be \textit{some observable} difference in the object-timestamps sets between the two snapshots.
This difference spans trivialities such as a minor edit in spelling to some object to significant differences such as the presence of new object-entities, e.g. consider the \textit{subject-property} pair \textit{John-sibling} with $O_{\mathcal{K}_{2019}} = \{\}$ and $O_{\mathcal{K}_{2020}} = \{\left(\textit{Maria},\_\_\right), \left(\textit{Anna}, \_\_\right), \left(\textit{Julian}, \_\_\right), \left(\textit{Roger}, \_\_\right)\}$\footnote{In these examples, we omit valid as well as transaction timestamps, denoted by `\_\_', since \textit{observable property change} does not employ them.}. This example would pass as having an \textit{observable property change}. It is, however, unlikely it reflects a real-world change, i.e., it is questionable whether \textit{John} had four new \textit{sibling}s in a single year, and is likely that the four objects were added as a delayed completion effort instead. Consider another example with \textit{subject-property} pair \textit{John-citizenship} with $O_{\mathcal{K}_{2019}} = \{\left(\textit{UK}, \_\_\right)\}$ and $O_{\mathcal{K}_{2020}} = \{\left(\textit{Britain}, \_\_\right)\}$. This would also pass as an \textit{observable property change} but would not be reflective of any real-world change, since the only difference is renaming \textit{UK} as \textit{Britain}, which is likely only an ontology refinement (correction). Hence, any observable property change subsumes three distinct artefacts - change due to real-world change, delayed completion, and correction. In this work, we are interested in changes due to real-world changes and stability thereof. 

\subsection{Exploring KB changes}

In order to get a better understanding of the types of changes, we consider changes that \textit{Ronaldo}'s $42$ properties - ranging from \textit{gender}, \textit{member of sports team} to \textit{official website} and \textit{social media followers}\footnote{We choose this set based on the current set of properties that entity \textit{Ronaldo} possesses in Wikidata, as of 28 January, 2021 - \url{https://www.wikidata.org/w/index.php?title=Q11571&oldid=1350069233}.} - experienced in an example year, $2018$. In total, the properties experienced a total of $27$ \textit{observable property changes}. Upon manual inspection, we observed --- only $2$ out of $27$ changes corresponded to changes due on real-world change, e.g., addition of object \textit{Juventus} to property \textit{member of sports team}; $10$ due to delayed completion, e.g., addition to the property \textit{father}; remaining $15$ due to correction, e.g., from $185\text{cm}$ to $185.4\text{cm}$ for the property \textit{height}.

\paragraph{Broader analysis} 
We extended the same analysis to now consider the same set of properties for \num{2533} \textit{Premier League Footballers} in Wikidata for the year $2020$, and observed \num{4429} changes in total. Out of these changes, we chose a sample of \num{4} properties: \textit{member of sports team}, \textit{participated in}, \textit{country for sport}, and \textit{award won}, and sampled \num{50} changes in each, distributed uniformly randomly across all entities. Again, we manually inspected the reason behind each change and annotated each change. The distribution of the number of changes are shown in Table \ref{tab:change-dist}. \textit{We observe that only about 10\% of the changes are due to real-world change and the rest are due to post-hoc completions and corrections.}

\begin{table}[t]
\resizebox{\columnwidth}{!}{%
\begin{tabular}{@{}cccc@{}}
\toprule
\textbf{Property} & \textbf{Real-world evolution} & \textbf{Post-hoc completion} & \textbf{Correction} \\ \midrule
team         & 17\%       &  47\%       &   35\%     \\
award        & 15\%       &   58\%      &   26\%     \\
participated & 2\%       &    67\%     &  31\%      \\
country      & 0       & 76\%        &    24\%    \\ \midrule
total        & 9\% & 62\% & 29\% \\ \bottomrule
\end{tabular}%
}
\caption{Showing number of changes corresponding to each property as categorized into the three reasons for changes --- (i) real-world change, (ii) delayed completion, (iii) correction. We see that 90\% of all changes encountered are in fact due to post-hoc completions and corrections. Note - \textit{team} stands for \textit{member of sports team}, \textit{awards} stands for \textit{awards won}, \textit{country} stands for \textit{country for sport} and \textit{participated} stands for \textit{participated in}.}
\label{tab:change-dist}
\end{table}

\subsection{Identifying real-world based KB changes}

\paragraph{Heuristic criteria for real-world based change identification} We observe firstly, that correction based changes are easy to identify - they are essentially \textit{edits}, \textit{replacements} and/or \textit{removals} of existing values, and are hence easy to filter out. Subsequently, we are left with changes due to real-world evolution and post-hoc completions. In order to distinguish between the two, we propose three heuristic criteria. Following notation introduced earlier, we present three heuristic criteria to determine whether $p$ for $s$ between sampling timestamps $\tau_1$ and $\tau_2$ has experienced change due to real-world change:
\begin{enumerate}
    \item \textbf{Timestamp Criterion}: We can base our criterion on changes that come with an explicit timestamp in the relevant interval, that is: $\exists (o,t_v,\_) \in O_{\mathcal{K}_{\tau_2}}:\tau_1 < t_v \leq \tau_2$, e.g., consider \textit{entity-property} pair $(\textit{Ronaldo, team})$ and points in time $\tau_1 = 2017$ and $\tau_2 = 2020$ and in $\mathcal{K}_{2020}$ we have $(\textit{Juventus}, 2018)$, then we say \textit{team} for \textit{Ronaldo} has changed between $2017$ and $2020$ due to real-world change. 
    \item \textbf{PCA Criterion}: This is based on the partial completeness assumption \cite{galarraga2013amie} which states that all objects have been added for a subject-relation-pair, whenever at least one object for it is present. Formally: $\exists (o,\_\_) \in O_{\mathcal{K}_{\tau_2}}: (o,\_\_) \notin O_{\mathcal{K}_{\tau_1}}$ and $O_{\mathcal{K}_{\tau_1}} \neq \emptyset$, i.e., there must be a \textit{new object} added between $\tau_1$ and $\tau_2$ to the already existing \textit{non-empty} list of objects present at $\tau_1$. The latter condition follows the PCA oracle and thereby assumes that the non-empty list is in fact a \textit{complete list}, and the criterion then posits tautologically that adding something to an already complete list reflects a real-world change. 
    \item \textbf{Bulk Criterion}: Another idea is to consider bulk updates as indicators of delayed insertion. Formally, $\neg \forall o_1, o_2:t_{a}\left(o_1\right) = t_{a} \left(o_2\right)$ where $\{(o_1,\_,t_a\left(o_1\right)), (o_2,\_,t_a\left(o_2\right)\} \in O_{\mathcal{K}_{t_2}} \text{and }\{(o_1,\_,t_a\left(o_1\right)), (o_2,\_,t_a\left(o_2\right)\} \notin O_{\mathcal{K}_{t_1}} \text{and }o_1 \neq o_2$. That is, not all \textit{newly} added distinct objects must be added at the same time (in bulk), where we measure time in resolution of one day. This heuristic was inspired from the observation that many one-off bulk additions to a property were generally delayed completion efforts.
\end{enumerate}


\paragraph{Evaluating heuristic criteria for real-world based change identification} Using the same manually annotated data as in the broader analysis for the sample of changes, we evaluated the accuracy of our heuristics (Table \ref{tab:change-heuristics}). Our findings are as follows.

\begin{enumerate}
    \item \textbf{Bulk update criterion} shows discouraging performance scores. We found that about \num{44}\% of all the changes due to real-world change underwent bulk updates. Further inspection suggests that a real-world change may in fact trigger bulk updates.
    \item \textbf{PCA criterion} routinely overestimates the number of changes as due to real-world evolution (high recall, low precision). For properties as far as our data is concerned, PCA seems to be a poor completeness oracle.
    \item \textbf{Timestamp criterion} is a highly precise criterion to identify changes due to real-world change, which is quite an expected result. 
\end{enumerate}
    
Timestamp criterion fails, obviously, when property values do not consist of any valid timestamps but the property indeed experiences changes due to real-world change. Not having valid timestamps is the norm for most properties, e.g., in the \num{2533} \textit{Premier League Footballers}, the property with the highest number of timestamped values is \textit{member of sports team} with 99\% of entities having at least one timestamped value, which is followed  by \textit{number of matches played} but with a 3.9\% of entities having at least one timestamped value, and subsequently, the percentages drop further rapidly. On the upside, what we did not expect but observed, is that there are not many real-world evolution changes in any of the properties that do not have often have timestamps. For example, \textit{participated in} is one such property that does not have timestamps, and had only \textit{1} change due to real-world evolution, and \textit{country for sport} which is another such property without timestamps underwent no changes whatsoever. Hence, not only do we observe that the timestamp criterion is highly precise, but in practice is also high recall.

\begin{table}[t]
\centering
\resizebox{\columnwidth}{!}{%
\begin{tabular}{l|lll|lll|lll}
\hline
\textbf{Properties}                    & \multicolumn{3}{l|}{\textbf{Timestamp}} & \multicolumn{3}{l|}{\textbf{PCA}} & \multicolumn{3}{l}{\textbf{Bulk}} \\ \cline{2-10} 
                           & P      & R      & F1     & P        & R        & F1       & P      & R      & F1      \\ \hline
all &   0.83     &  0.87      & 0.84       &   0.23       &    0.87      &      0.36    &    0.08    &     0.44   &    0.14     \\ \hline
awards  &    1.00    &  0.71      &     0.83   &    0.27      &     0.71     &   0.40       &   0.20     &     0.41   &   0.27      \\ \hline
team   &    1.00    &    1.00    &    1.00    &     0.27     &      1.00    &  0.43        &    0.36    &   0.44     &      0.40   \\ \hline
\end{tabular}%
}
\caption{Table showing evaluation of heuristic criteria that identify whether a given change is due to real-world change or not. The properties column represents for what set of properties, the evaluation was performed on. Firstly, note that timestamp criterion provides promising performance. Secondly, also note that the performance of timestamp criterion is best when considering property \textit{team}, and subsequently gets worse when  other properties are considered together. This can be explained by the lack of timestamps for other properties such as \textit{country} and \textit{participated}. Note that `all' in Properties column refers to all four properties considered together.}
\label{tab:change-heuristics}
\end{table}




 


\section{Heuristics for Stability}



Now that we understand, heuristically, how to identify changes due to real-world change, we approach the problem of predicting property change (as formalized in Section \ref{sec:formalization}). First, we notice that some entities, e.g., \textit{dead} \textit{Humans}, and some properties, e.g., \textit{birthplace} are inherently stable. Before tackling the problem of predicting property change, we first consider filtering away these inherently stable entities and properties, and in this section present methods thereof.

\subsection{Stable entities}

We investigated two approaches - based on structured features (having a end date, e.g., death date for humans, dissolution date for companies etc.), and on heuristics from Wikipedia edit activity and page growth (as an indicator of true change, compared to minor edits).
 




\subsection{Stable properties}

Regardless of stability in entities, there exists only a subset of properties of the entities that change due to real-world change, e.g., \textit{residence} changes and \textit{birthplace} does not. We filter properties that are not the kind that change (stable properties). The idea that we use in order to classify given a property as one that is inherently unstable is the following \textit{filtering heuristic} - \textit{``if a property has changed at least once, for 5\% of the total number of entities (of a class) in the past then it's an unstable property"}. In order to estimate the number of times a property has changed since the beginning of time, we present the following heuristic measures.

\begin{enumerate}
    \item \textbf{KB historic edits}: The first one we consider is based on the total number of times the property has undergone edits in the past measured via access to the historical revisions to the KB. The measured change is the same as \textit{observable property change}. \vspace{1mm} \\
    \item \textbf{Multiplicity in property values}: We then turn to a much simpler method, using the multiplicity in the number of objects of the property of an entity, read using the current KB snapshot, i.e., we consider the number of objects that a property for an entity contains to be equal to the number of times the property of the entity has changed due to real-world change. \vspace{1mm}\\ 
    \item \textbf{Multiplicity in timestamped property values}: Similar to previous, here we consider the number of \textit{timestamped} objects that a property of entity contains to be equal to the number of times the property of the entity has changed due to real-world change. 
\end{enumerate}


In order to evaluate the measures with the \textit{filtering heuristic}, we consider multiple large random samples of entities, i.e., \num{50000} \textit{Politicians}, \num{30000} \textit{Football Players}, \num{30000} \textit{Actors}, \num{3916} \textit{Enterprises} and \num{5614} \textit{Music Groups}, and for each sample set, we consider the most popular and interesting properties (about 15 of them in each case). We apply the filter for the properties separately for each sample set of entities, and classify the properties into stable and unstable ones using the aforementioned filtering technique (1 - stable, 0 - unstable). In parallel, we also manually classify these properties in each case into stable and unstable ones relying on real-world intuition, and use them as ground-truth in order to evaluate the performance of the classification via the filter, and report the performance in Table \ref{tab:property-filtering}.

\begin{table}[t]
\centering
\resizebox{\columnwidth}{!}{%
\begin{tabular}{|c|c|c|c|c|c|c|c|c|c|}
\hline
\textbf{Class} & \multicolumn{3}{c|}{\textbf{KB edits}} & \multicolumn{3}{c|}{\textbf{Object multiplicity}} & \multicolumn{3}{c|}{\textbf{Timestamp multiplicity}} \\ \cline{2-10} 
               & \textbf{P}  & \textbf{R} & \textbf{F1} & \textbf{P}   & \textbf{R}   & \textbf{F1}  & \textbf{P}  & \textbf{R}  & \textbf{F1}  \\ \hline
Footballers  & 0    & 0    & 0    & 0.80 & 1.00 & 0.88 & 1.00 & 0.40 & 0.57 \\ \hline
Politicians  & 1.00 & 0.25 & 0.40 & 0.82 & 1.00 & 0.70 & 1.00 & 0.61 & 0.77 \\ \hline
Actors       & 0.70    & 0.20    & 0.31    & 0.80 & 0.88 & 0.84 & 1.00 & 0.54 & 0.70 \\ \hline
Enterprises  &  0.40    &  0.30    &   0.34   & 0.86 & 0.95 & 0.90 & 1.00 & 0.74 & 0.85 \\ \hline
Music groups &  0.60    &   0.40   &   0.48   & 0.83 & 1.00 & 0.91 & 1.00 & 0.67 & 0.80 \\ \hline
\end{tabular}%
}
\caption{Table showing results for property filtering heuristics (for changing properties).}
\label{tab:property-filtering}
\end{table}






\section{Stability as Supervised Prediction Problem} \label{section:supervised-prediction}


After filtering out stable entities and stable properties, we consider the core problem of property change prediction, which we treat as a \textit{supervised binary classification problem}, i.e., given a \textit{subject-property} pair $\left(s,p\right)$, we predict whether $p$ would change as a result of real-world change between sampling timepoints $\tau_1$ and $\tau_2$, where $\tau_2 > \tau_1$, using various features of $s$. The prediction target, given the features of $s$, is a binary variable that denotes whether $p$ of $s$ between $\tau_1$ and $\tau_2$ or not.

\subsection{Features}

We consider the following orthogonal classes of features of $s$ that we posit to be predictive of change of $p$:

\paragraph{Observable KB features}  This class posits that the properties and corresponding objects of $s$, other than $p$, are predictive of change of $p$, e.g., \textit{age} of a \textit{Footballer} is predictive of change of \textit{member of sports team}. There are two challenges in generalizing this to all the properties --- (i) most property values, unlike \textit{age}, are not literal numerical values, but are either literal strings, such as names, or specific entities, such as a country in the KB, hence representing them to use as a feature is a challenge, and (ii) some properties may not be present for all entities to use, e.g., not all \textit{Footballers} might have the property \textit{positions played}. One way in which we overcome these two challenges is by considering all properties and corresponding objects that an entity has as Bag-of-Words (\textit{BoW}) unigram text and use either count- or \textit{tf-idf}-based vectorization to employ as features (denoted as \textit{structured-bow}).

\paragraph{Latent KB features} Another solution to the problem of observable KB features is to alternatively use an alternative entity representation such as entity's \textit{KB embedding}. Not only are KB embeddings numerical and general across all classes in a KB, they are also shown to capture latent features of entities such as `how famous', `how central' an entity is~\cite{nickel2015review} and we posit that these features are very predictive in property change.

\paragraph{Wikipedia page text features} One could also use an external source like Wikipedia of $s$ (when available) in order to predict property change, e.g., words such as ``injured" or ``retired" in a \textit{Footballer}'s Wikipedia article suggests that the property \textit{member of sports team} is likely to not change. One could either use the latest version of the Wikipedia page, i.e., version at $\tau_2$, or consider only the text that is modified or added in between $\tau_1$ and $\tau_2$. 
Once that is decided, we convert the text into \textit{BoW}, and derive \textit{unigram}, \textit{bigram} as well as \textit{trigram} features, and use a standard \textit{tf-idf} weighting scheme.

\paragraph{Graph neighborhood features}   Another idea is that given a property of an entity, whose stability is to be estimated between $\tau_1$ and $\tau_2$, we look at how similar entities have behaved between $\tau_1$ and $\tau_2$ with respect to the same property. We represent entities via KB embeddings and consider \textit{K}-nearest-neighbors of the entity, computed via Euclidean distance, and take the proportion of those neighbors for which the property has changed between $\tau_1$ and $\tau_2$ as the feature.

\subsection{Model}

Using each of the features as described and the aforementioned target, we train a logistic regression model, \textit{per property per class}. For example, consider property \textit{award won} ($p$) of the sample of \textit{Actor}s, $S = \{s_i\}_{i=1}^{n}, n \in \mathbb{N} \cup\{0\}$. We choose a sampling period $(\tau_1, \tau_2] = (2017, 2020]$ across all experiments. Then, for each \textit{subject-property} $(s_i,p)$, we check whether $p$ has changed in $(2017, 2020]$ and set corresponding target, $y_{p, s_i}$ as \num{1} if changed, \num{0} otherwise. 
We then extract the interested feature of $s_i$ that we generically denote as $X_{s_i}$, e.g., text unigram features using $s$'s current Wikipedia page with tf-idf weighting. Once we do this for all of $S$ then we have our dataset $D = {\left\{\left(X_{s_i}, y_{p, s_i}\right)\right\}_{i=1}^{n}}$.

\begin{table*}[t]
\centering
\resizebox{\textwidth}{!}{%
\begin{tabular}{|c|c|c|c|c|c|c|c|c|c|c|c|c|}
\hline
\textbf{Feature} &
  \multicolumn{3}{c|}{\textbf{\begin{tabular}[c]{@{}c@{}}member of sports team\\ Football players\end{tabular}}} &
  \multicolumn{3}{c|}{\textbf{\begin{tabular}[c]{@{}c@{}}positions held\\ Politicians\end{tabular}}} &
  \multicolumn{3}{c|}{\textbf{\begin{tabular}[c]{@{}c@{}}awards won\\ Actors\end{tabular}}} &
  \multicolumn{3}{c|}{\textbf{\begin{tabular}[c]{@{}c@{}}head coach\\ Football clubs\end{tabular}}} \\ \cline{2-13} 
\textbf{} &
  \textbf{P} &
  \textbf{R} &
  \textbf{F1} &
  \textbf{P} &
  \textbf{R} &
  \textbf{F1} &
  \textbf{P} &
  \textbf{R} &
  \textbf{F1} &
  \textbf{P} &
  \textbf{R} &
  \textbf{F1} \\ \hline
\textbf{\begin{tabular}[c]{@{}c@{}}Wikipedia text\\\end{tabular}} &
  0.81 &
  0.85 &
  0.83 &
  0.75 &
  0.72 &
  0.73 &
  0.70 &
  0.73 &
  0.72 &
  0.61 &
  0.68 &
  0.64 \\ \hline
\textbf{\begin{tabular}[c]{@{}c@{}}Age\\\end{tabular}} &
  0.71 &
  0.85 &
  0.75 &
  0.56 &
  0.50 &
  0.53 &
  0.53 &
  0.56 &
  0.54 &
  - &
  - &
  - \\ \hline
\textbf{\begin{tabular}[c]{@{}c@{}}Embedding\\ \end{tabular}} &
  0.79 &
  0.80 &
  0.79 &
  0.64 &
  0.68 &
  0.66 &
  0.63 &
  0.61 &
  0.62 &
  0.54 &
  0.64 &
  0.58 \\ \hline
\textbf{\begin{tabular}[c]{@{}c@{}}Structured-BoW\end{tabular}} &
  0.77 &
  0.74 &
  0.76 &
  0.66 &
  0.67 &
  0.66 &
  0.53 &
  0.76 &
  0.62 &
  0.60 &
  0.61 &
  0.60 \\ \hline
\textbf{\begin{tabular}[c]{@{}c@{}}Graph neighborhood\\ (using kNN)\end{tabular}} &
  0.74 &
  0.73 &
  0.74 &
  0.68 &
  0.68 &
  0.68 &
  0.66 &
  0.73 &
  0.70 &
  0.60 &
  0.59 &
  0.60 \\ \hline
\end{tabular}%
}
\caption{Test performance (P: precision, R: recall, F1: F1 score) for different properties. For each cell, assume random baseline of \num{50}\%. 
}
\label{tab:prop-change-results}
\end{table*}

\subsection{Evaluation}

\paragraph{Experiments} In order to evaluate the ideas for supervised prediction, we consider the following different unstable properties along with respective class --- (i) affiliation type: \textit{member of sports team} for \textit{Footballers} and \textit{Basketball players}, \textit{positions held} for \textit{Politicians}, (ii) award type: \textit{award won} as well as \textit{nominations} for \textit{Footballers}, \textit{Basketball players}, \textit{Actors} and \textit{Politicians}, (iii) head of organization type: \textit{head coach} of \textit{Football clubs}. 
For each of these properties and the respective classes of entities we train logistic regression models considering each feature individually first and then all together, with target being \textit{change} ($y_{p, s_i} = \num{1}$) / \textit{no-change} ($y_{p, s_i} = 0$) measured using timestamp criterion for sampling period $(2017,2020]$. For each of the model evaluations, we balance the datasets for training as well as testing and hence assume a \num{50}\% baseline performance (reported as precision \textit{p}, recall \textit{r}, F1-score \textit{f1}) across all models, and we dedicate a conservative minimum of \num{40}\% of the data for the test set, and report performance on the test set in Table \ref{tab:prop-change-results}.

\paragraph{Results} We observe that the highest performing feature, consistently across all properties, is Wikipedia text based, specifically, unigram features from the latest Wikipedia article of the entity vectorized via tf-idf weighting. We also see that almost all results cross the \num{50}\% random baseline, supporting the arguments for the use of the features discussed.

\paragraph{Model inspection} In order to get an insight into the models, we inspected the features learned in each of the cases. (i) When trained using text features, we observe that the model picks up a lot of the expected unigrams, e.g., for \textit{member of sports team}, we see that many of the positively weighted features are words such as ``new", ``club", ``contract", ``sign", signifying \textit{Footballers}' team change and analogously negatively weighted features contain ``manager", ``career", ``coach" signifying \textit{Footballers}' retirement from being a player; similarly for \textit{positions held} of \textit{Politicians}, positive features contain ``elected", ``vote", ``office"; and positive features for \textit{award won} contain ``best", ``performance", ``received", ``award" and so on. Furthermore, in each case, dates in text lying in $(2017,2020]$ such as ``2018" and ``2019" are found to be positively weighted and dates far in the past such as ``2007", ``2011" and so on are negatively weighted signifying the use recency of events for each entity leading to change / no-change. But, we do note that there do exist a majority of features that are not interpretable, resulting from spurious correlations in the data. (ii) When it comes to using to structured features, we firstly observe as we expected, that \textit{age} is only good predictor in selected cases such as predicting change of \textit{member of sports team} where the model predicts \textit{change} for younger players ($<30$ years) and achieves reasonably high performance that is interpretable, but a poor one for to all other cases and supports the argument that using specific property values can only be used in specific cases and does not generalize. Using \textit{structured-bow} features, we see that the results are encouraging, but upon inspecting the unigrams learned, we see hardly any interpretable ones, denoting again, the model picking up spurious correlations in the data-generation process, e.g., positive weights for unigrams such as ``Italy", ``USA" and so on. (iii) When it comes to using KB embedding as a feature for logistic regression, the results are highly encouraging, except that KB embeddings represent latent features of entities in the KB and are not human-interpretable. 

\paragraph{Manual Evaluation} Given time period \textit{2017--2020}, we took random balanced set of 50 \textit{Footballers}, half of whom had changed \textit{member of sports team} and half did not (automatically annotated via Timestamp criterion). We then proceeded to predict property change manually for each entity, using the entity's Wikipedia page, entity's age and just general intuition where possible. Our manual annotation achieved an $\sim0.80$ \textit{f1} (similar \textit{p} and \textit{r}). We learn that extracting change-event information in Wikipedia article provides the most accurate way to perform the prediction (such as ``\textit{X} moved from \textit{Barcelona} to \textit{Real Madrid} in \textit{2018}"), followed by clues in the article that imply whether or not the entity as a whole or a property \textit{can} change (such as, ``\textit{X} is retired"). However, in cases where the text does not contain explicit change-event information or clues, it does not necessarily mean that the property of the entity did not undergo change, as the text article is itself incomplete or completely absent. In such cases, we made guesses of completeness based on the length of the Wikipedia article and if the article is long enough and does not yet contain change-event information or clues then we predict that the property of the entity did not change. We also used \textit{age} of the entity when Wikipedia article was not informative and also the last time the property for the entity has changed and make a prediction based on that, which ultimately amount more towards guessing than rigorously explicable predictions, illuminating the difficulty of the task at hand.










\section{Discussion}
Having presented all the ideas to table the problems of KB stability, we now critically reflect on the work and provide pointers for future work.


\paragraph{Ground Truth for Supervised Prediction} In order to perform supervised classification of change / no change of properties of entities, we require to acquired ground truths for change / no-change information in datasets using the timestamp criterion. The gold standard for such ground truth is generally human annotation, which is often too expensive and time consuming, and in our case non-trivial to do correctly since it is unclear what ground-truth source must be used that has information on events, if the entities are long tail. Resorting to using our high-precision heuristic timestamp criterion a tremendous advantage that it is automatic and high-precision, but a disadvantage that it is potentially low recall, since some entities in the KB may not have been updated with the required event, and even if the event has been updated, it may not contain a timestamp, which is often a problem with crowd-curated KBs, which is a source of noise in our datasets.



\paragraph{Change Time Distribution Estimation} One can also tackle the change prediction problem by approaching it as a inter-event time distribution estimation problem, thereby specifically addressing temporal dynamics. That is, if we consider change to a property as a random event in time and we consider several changes to the same property, then what we get is a empirical set of inter-event times, e.g., consider the list of objects with timestamps for \textit{Ronaldo}'s \textit{member of sports team} - [(\textit{Sporting CP, start:2002}),  (\textit{ManU, start:2003}), (\textit{Real Madrid, start:2009}), (\textit{Juventus, start:2018})], and corresponding inter-change times - [1, 6, 9] (found by subtracting start-times in years). We can then hypothesize that this empirical set of change-event times has underlying change-event time distribution. Access to this true distribution allows us to precisely answer questions such as, ``when next would the property change, given its history?". 

\begin{figure}[t]
    \centering
    \includegraphics[scale=0.24]{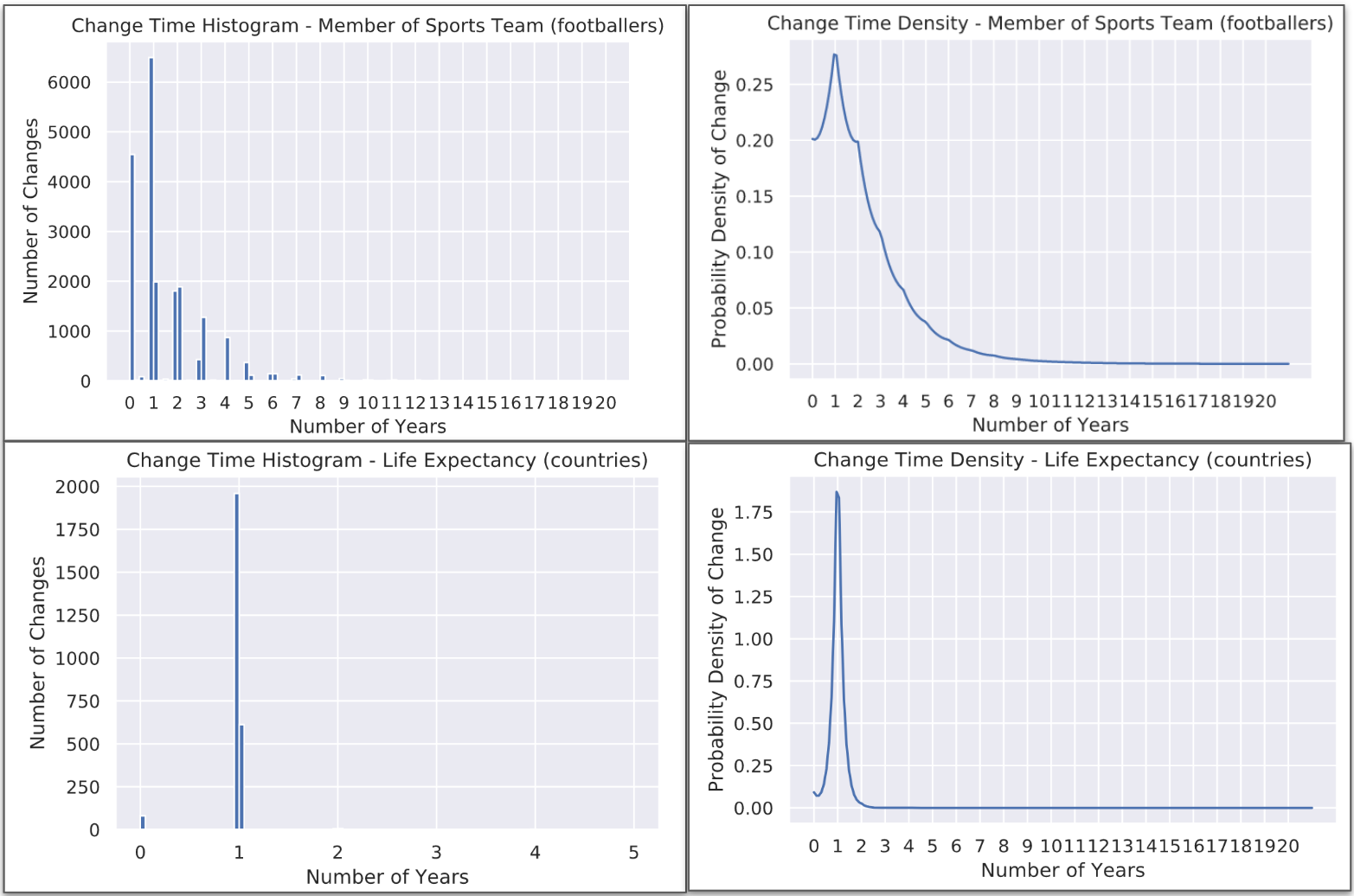}
    \caption{Figure showing, on the leftmost column, histograms of inter-change times for set of \num{20756} changes to property \textit{member of sports team} sampled from a random subset of \num{5000} \textit{Footballers} (above) and for a set of \num{2667} inter-change times for \textit{life expectancy} for a total of \num{170} countries in Wikidata (below). The plots in the rightmost column show corresponding smooth approximation of the histogram to obtain likelihoods using KDE.}
    \label{fig:kde}
\end{figure}

\paragraph{KDE} At first, we can make the most simplifying assumption about the change-events --- that they are independent of each other, and hence so are the inter-change times. Considering such inter-change times from \num{5000} \textit{Footballers} in Wikidata, we plotted a histogram of (assumed to be independent) inter-change times, and consequently, in order to get a smooth approximation of the histogram, we use \textit{kernel density estimation} (KDE) (Fig \ref{fig:kde}). Such information alongside property values is essentially stability information and can help users as well as maintainers understand the change-likelihood based on a large sample of changes to the specific property. We find the use of KDE as an especially attractive choice for understanding change-likelihood in periodic properties such \textit{life expectancy} (Figure \ref{fig:kde} or \textit{GDP} of a country, \textit{net profit} or \textit{total revenue} of an enterprise, and so on. Some of these periodic properties represent a special case, in that they are in the real-world \textit{continuously} changing (as opposed to discrete, such as changing a team), but only sampled in discrete points in time like \textit{life expectancy} is sampled once a year, \textit{net profit} every quarter, and so on. Another unique characteristic of periodically changing properties in general is that the change-events may hardly correlate with other characteristics of the entity that it is part of, i.e., these properties get updated regardless of other features of the entity such as the content of its Wikipedia page, or explicit characteristics such as \textit{age}.

\paragraph{Limitations of KDE} Naturally though, KDE and histograms make a strong \textit{i.i.d} assumption, i.e., the likelihoods w.r.t time that they estimate is simply based on the frequencies of inter-change times. In reality, these changes are hardly independent of each other. Changes to any property are in fact sequential, with \textit{short-} and \textit{long-range} dependencies among one another. Consider again, \textit{member of sports team} of \textit{Footballers}. A repetitive pattern that one can observe is that \textit{Footballers} in the beginning of their careers change \textit{member of sports team} every year and towards the latter half change only once in 3-4 years, until eventual retirement. Secondly, this temporal trend does depend on factors such as if the \textit{Footballer} is injured in between or is prematurely retired and so on, and is not an independently temporally changing property such as periodic properties, which are ignored by KDE and histograms. This requires that we employ sophisticated temporal models such as \textit{temporal point processes}~\cite{rasmussen2018lecture}, especially those that allow arbitrary trends to be captured via neural networks~\cite{shchur2019intensity, omi2019fully, upadhyay2018deep}.

\subsection{What is Fundamentally Stable?}

\paragraph{Entities that undergo real-world change} The problem of stability due to real-world evolution does not apply to a KB throughout, since KBs often contain certain \textit{timeless} entities or entity classes even such as the class of \textit{color}s\footnote{\url{https://www.wikidata.org/wiki/Q1075}} or the entity of color \textit{blue}\footnote{\url{https://www.wikidata.org/wiki/Q1088}}. In Wikidata, in general, there are large parts of its ontology that are virtually fully stable such as the class of \textit{chemical compounds} which consists of over a million different chemical compounds as individual entities, accounting for $1.7\%$ of total entity count; class of \textit{taxon}, which consists of nearly 3 million different taxonomic groups of various ranks such as a species, families, or classes of organisms, accounting for $3.8\%$ of total entity count, and so on. Such classes and groups of entities can be identified using the knowledge of the content of a KB. This helps in trusting the expected temporal validity of statements in a potentially large part of the KB and also reduces the burden of \textit{timeliness}-based update for the same part of the KB. Specifically, entities that undergo real-world evolution based changes are typically leaf-level \textit{instances} or \textit{named entities} that have a practically limited lifespan such as \textit{human}s. 

\paragraph{Real-world evolution stability is only a problem for real-world KBs} Lastly, real-world evolution based stability is not a problem for certain kinds of KBs that are static by design. Commonsense KBs are an illustrative of the same, which rather contain only general \textit{concepts} like \textit{eating breakfast} as entities instead of concrete \textit{named entities} like \textit{Ronaldo}. For example, commonsense KB ATOMIC (\cite{atomic}) consists of events connected by \textit{if-then} relationships, such as the entity `\textit{X repels Y's attacks}' is connected via relationship `\textit{X is seen as}' to entity `\textit{X is strong}'. Similarly, a causal KB, one that consists of causal-effect concepts closely related to commonsense KBs, like CauseNet (\cite{heindorf2020causenet}) with entities such as `\textit{disease}' is connected via relationship `\textit{causes}' to entity `\textit{death}'. In such KBs, stability resulting from real-world evolution is not a problem.

\bibliographystyle{plain}          
\bibliography{bibliography}

\begin{thebibliography}{10}

\bibitem{abujabal2017quint}
Abdalghani Abujabal, Rishiraj~Saha Roy, Mohamed Yahya, and Gerhard Weikum.
\newblock Quint: Interpretable question answering over knowledge bases.
\newblock In {\em Proceedings of the 2017 Conference on Empirical Methods in
  Natural Language Processing: System Demonstrations}, pages 61--66, 2017.

\bibitem{almquist2019towards}
Axel Almquist and Adam Jatowt.
\newblock Towards content expiry date determination: Predicting validity
  periods of sentences.
\newblock In {\em ECIR}, 2019.

\bibitem{arnaout2021negative}
Hiba Arnaout, Simon Razniewski, Gerhard Weikum, and Jeff~Z Pan.
\newblock Negative statements considered useful.
\newblock {\em JWS}, 2021.

\bibitem{dbpedia}
S{\"o}ren Auer, Christian Bizer, Georgi Kobilarov, Jens Lehmann, Richard
  Cyganiak, and Zachary Ives.
\newblock Dbpedia: A nucleus for a web of open data.
\newblock In {\em The semantic web}, pages 722--735. Springer, 2007.

\bibitem{balaraman2018recoin}
Vevake Balaraman, Simon Razniewski, and Werner Nutt.
\newblock Recoin: relative completeness in wikidata.
\newblock In {\em Companion Proceedings of the The Web Conference 2018}, pages
  1787--1792, 2018.

\bibitem{bordes2013translating}
Antoine Bordes, Nicolas Usunier, Alberto Garcia-Duran, Jason Weston, and Oksana
  Yakhnenko.
\newblock Translating embeddings for modeling multi-relational data.
\newblock In {\em Neural Information Processing Systems (NIPS)}, pages 1--9,
  2013.

\bibitem{change-cho-molina}
Junghoo Cho and Hector Garcia-Molina.
\newblock Estimating frequency of change.
\newblock {\em ACM Trans. Internet Technol.}, 3(3):256–290, August 2003.

\bibitem{diefenbach2019qanswer}
Dennis Diefenbach, Pedro~Henrique Migliatti, Omar Qawasmeh, Vincent Lully,
  Kamal Singh, and Pierre Maret.
\newblock {QAnswer}: a question answering prototype bridging the gap between a
  considerable part of the lod cloud and end-users.
\newblock In {\em WWW}, 2019.

\bibitem{dikeoulias2019epitaph}
Ioannis Dikeoulias, Jannik Str{\"o}tgen, and Simon Razniewski.
\newblock Epitaph or breaking news? analyzing and predicting the stability of
  knowledge base properties.
\newblock In {\em Companion Proceedings of The 2019 World Wide Web Conference},
  pages 1155--1158, 2019.

\bibitem{dong2014knowledge}
Xin Dong, Evgeniy Gabrilovich, Geremy Heitz, Wilko Horn, Ni~Lao, Kevin Murphy,
  Thomas Strohmann, Shaohua Sun, and Wei Zhang.
\newblock Knowledge vault: A web-scale approach to probabilistic knowledge
  fusion.
\newblock In {\em Proceedings of the 20th ACM SIGKDD international conference
  on Knowledge discovery and data mining}, pages 601--610, 2014.

\bibitem{dubey2019lc}
Mohnish Dubey, Debayan Banerjee, Abdelrahman Abdelkawi, and Jens Lehmann.
\newblock Lc-quad 2.0: A large dataset for complex question answering over
  wikidata and dbpedia.
\newblock In {\em International semantic web conference}, pages 69--78, 2019.

\bibitem{galarraga2016completeness}
Luis Gal\'{a}rraga, Simon Razniewski, Antoine Amarilli, and Fabian~M. Suchanek.
\newblock Predicting completeness in knowledge bases.
\newblock In {\em Proceedings of the Tenth ACM International Conference on Web
  Search and Data Mining}, WSDM ’17, page 375–383, New York, NY, USA, 2017.
  Association for Computing Machinery.

\bibitem{galarraga2017predicting}
Luis Gal{\'a}rraga, Simon Razniewski, Antoine Amarilli, and Fabian~M Suchanek.
\newblock Predicting completeness in knowledge bases.
\newblock In {\em Proceedings of the tenth acm international conference on web
  search and data mining}, pages 375--383, 2017.

\bibitem{galarraga2013amie}
Luis~Antonio Gal{\'a}rraga, Christina Teflioudi, Katja Hose, and Fabian
  Suchanek.
\newblock Amie: association rule mining under incomplete evidence in
  ontological knowledge bases.
\newblock In {\em Proceedings of the 22nd international conference on World
  Wide Web}, pages 413--422, 2013.

\bibitem{gerber2015defacto}
Daniel Gerber, Diego Esteves, Jens Lehmann, Lorenz B{\"u}hmann, Ricardo Usbeck,
  Axel-Cyrille~Ngonga Ngomo, and Ren{\'e} Speck.
\newblock Defacto—temporal and multilingual deep fact validation.
\newblock {\em Journal of Web Semantics}, 35:85--101, 2015.

\bibitem{han2020graph}
Zhen Han, Yunpu Ma, Yuyi Wang, Stephan Guenemman, and Volker Tresp.
\newblock Graph hawkes neural network for forecasting on temporal knowledge
  graphs.
\newblock In {\em Automated Knowledge Base Construction}, 2020.

\bibitem{hao2020outdated}
Shuang Hao, Chengliang Chai, Guoliang Li, Nan Tang, Ning Wang, and Xiang Yu.
\newblock Outdated fact detection in knowledge bases.
\newblock In {\em 2020 IEEE 36th International Conference on Data Engineering
  (ICDE)}, pages 1890--1893. IEEE, 2020.

\bibitem{wikidata-vandalism}
Stefan Heindorf, Martin Potthast, Benno Stein, and Gregor Engels.
\newblock Vandalism detection in wikidata.
\newblock In {\em Proceedings of the 25th ACM International on Conference on
  Information and Knowledge Management}, CIKM '16, page 327–336, New York,
  NY, USA, 2016. Association for Computing Machinery.

\bibitem{heindorf2020causenet}
Stefan Heindorf, Yan Scholten, Henning Wachsmuth, Axel-Cyrille Ngonga~Ngomo,
  and Martin Potthast.
\newblock Causenet: Towards a causality graph extracted from the web.
\newblock In {\em Proceedings of the 29th ACM International Conference on
  Information \& Knowledge Management}, pages 3023--3030, 2020.

\bibitem{hogan2020knowledge}
Aidan Hogan, Eva Blomqvist, Michael Cochez, Claudia d'Amato, Gerard de~Melo,
  Claudio Gutierrez, Jos{\'e} Emilio~Labra Gayo, Sabrina Kirrane, Sebastian
  Neumaier, Axel Polleres, et~al.
\newblock Knowledge graphs.
\newblock {\em arXiv preprint arXiv:2003.02320}, 2020.

\bibitem{jiang2016towards}
Tingsong Jiang, Tianyu Liu, Tao Ge, Lei Sha, Baobao Chang, Sujian Li, and
  Zhifang Sui.
\newblock Towards time-aware knowledge graph completion.
\newblock In {\em Proceedings of COLING 2016, the 26th International Conference
  on Computational Linguistics: Technical Papers}, pages 1715--1724, 2016.

\bibitem{kuzey2016time}
Erdal Kuzey, Vinay Setty, Jannik Str{\"o}tgen, and Gerhard Weikum.
\newblock As time goes by: Comprehensive tagging of textual phrases with
  temporal scopes.
\newblock In {\em Proceedings of the 25th international conference on world
  wide web}, pages 915--925, 2016.

\bibitem{leblay2018deriving}
Julien Leblay and Melisachew~Wudage Chekol.
\newblock Deriving validity time in knowledge graph.
\newblock In {\em Companion Proceedings of the The Web Conference 2018}, pages
  1771--1776, 2018.

\bibitem{lehmann2010ore}
Jens Lehmann and Lorenz B{\"u}hmann.
\newblock Ore-a tool for repairing and enriching knowledge bases.
\newblock In {\em International Semantic Web Conference}, pages 177--193.
  Springer, 2010.

\bibitem{li2015probabilistic}
Huiying Li, Yuanyuan Li, Feifei Xu, and Xinyu Zhong.
\newblock Probabilistic error detecting in numerical linked data.
\newblock In {\em Database and Expert Systems Applications}, pages 61--75.
  Springer, 2015.

\bibitem{chinese-kb}
Jiaqing Liang, Sheng Zhang, and Yanghua Xiao134.
\newblock How to keep a knowledge base synchronized with its encyclopedia
  source.
\newblock In {\em IJCAI}, 2017.

\bibitem{luggen-completeness}
Michael Luggen, Djellel Difallah, Cristina Sarasua, Gianluca Demartini, and
  Philippe Cudr{\'e}-Mauroux.
\newblock Non-parametric class completeness estimators for collaborative
  knowledge graphs—the case of wikidata.
\newblock In {\em International Semantic Web Conference}, pages 453--469.
  Springer, 2019.

\bibitem{ma2014learning}
Yanfang Ma, Huan Gao, Tianxing Wu, and Guilin Qi.
\newblock Learning disjointness axioms with association rule mining and its
  application to inconsistency detection of linked data.
\newblock In {\em Chinese Semantic Web and Web Science Conference}, pages
  29--41. Springer, 2014.

\bibitem{ma2019embedding}
Yunpu Ma, Volker Tresp, and Erik~A Daxberger.
\newblock Embedding models for episodic knowledge graphs.
\newblock {\em Journal of Web Semantics}, 59:100490, 2019.

\bibitem{nickel2015review}
Maximilian Nickel, Kevin Murphy, Volker Tresp, and Evgeniy Gabrilovich.
\newblock A review of relational machine learning for knowledge graphs.
\newblock {\em Proceedings of the IEEE}, 104(1):11--33, 2015.

\bibitem{factorize-YAGO}
Maximilian Nickel, Volker Tresp, and Hans-Peter Kriegel.
\newblock Factorizing yago: Scalable machine learning for linked data.
\newblock In {\em Proceedings of the 21st International Conference on World
  Wide Web}, WWW '12, page 271–280, New York, NY, USA, 2012. Association for
  Computing Machinery.

\bibitem{nickel2012factorizing}
Maximilian Nickel, Volker Tresp, and Hans-Peter Kriegel.
\newblock Factorizing yago: scalable machine learning for linked data.
\newblock In {\em Proceedings of the 21st international conference on World
  Wide Web}, pages 271--280, 2012.

\bibitem{industry-kg}
Natasha Noy, Yuqing Gao, Anshu Jain, Anant Narayanan, Alan Patterson, and Jamie
  Taylor.
\newblock Industry-scale knowledge graphs: Lessons and challenges.
\newblock {\em Commun. ACM}, 62(8):36–43, July 2019.

\bibitem{omi2019fully}
Takahiro Omi, Naonori Ueda, and Kazuyuki Aihara.
\newblock Fully neural network based model for general temporal point
  processes.
\newblock {\em arXiv preprint arXiv:1905.09690}, 2019.

\bibitem{paulheim2017knowledge}
Heiko Paulheim.
\newblock Knowledge graph refinement: A survey of approaches and evaluation
  methods.
\newblock {\em Semantic web}, 8(3):489--508, 2017.

\bibitem{paulheim2014improving}
Heiko Paulheim and Christian Bizer.
\newblock Improving the quality of linked data using statistical distributions.
\newblock {\em International Journal on Semantic Web and Information Systems
  (IJSWIS)}, 10(2):63--86, 2014.

\bibitem{pipino2002quality}
Leo~L. Pipino, Yang~W. Lee, and Richard~Y. Wang.
\newblock Data quality assessment.
\newblock {\em Commun. ACM}, 45(4):211–218, April 2002.

\bibitem{rashid2019quality}
Mohammad Rashid, Marco Torchiano, Giuseppe Rizzo, Nandana Mihindukulasooriya,
  and Oscar Corcho.
\newblock A quality assessment approach for evolving knowledge bases.
\newblock {\em Semantic Web}, 10(2):349--383, 2019.

\bibitem{rasmussen2018lecture}
Jakob~Gulddahl Rasmussen.
\newblock Lecture notes: Temporal point processes and the conditional intensity
  function.
\newblock {\em arXiv preprint arXiv:1806.00221}, 2018.

\bibitem{razniewski2016optimizing}
Simon Razniewski.
\newblock Optimizing update frequencies for decaying information.
\newblock In {\em Proceedings of the 25th ACM International on Conference on
  Information and Knowledge Management}, pages 1191--1200, 2016.

\bibitem{atomic}
Maarten Sap, Ronan Le~Bras, Emily Allaway, Chandra Bhagavatula, Nicholas
  Lourie, Hannah Rashkin, Brendan Roof, Noah~A Smith, and Yejin Choi.
\newblock Atomic: An atlas of machine commonsense for if-then reasoning.
\newblock In {\em Proceedings of the AAAI conference on artificial
  intelligence}, 2019.

\bibitem{shchur2019intensity}
Oleksandr Shchur, Marin Bilo{\v{s}}, and Stephan G{\"u}nnemann.
\newblock Intensity-free learning of temporal point processes.
\newblock {\em arXiv preprint arXiv:1909.12127}, 2019.

\bibitem{socher2013reasoning}
Richard Socher, Danqi Chen, Christopher~D Manning, and Andrew Ng.
\newblock Reasoning with neural tensor networks for knowledge base completion.
\newblock In {\em Advances in neural information processing systems}, pages
  926--934. Citeseer, 2013.

\bibitem{YAGO}
Fabian~M. Suchanek, Gjergji Kasneci, and Gerhard Weikum.
\newblock Yago: A core of semantic knowledge.
\newblock In {\em Proceedings of the 16th International Conference on World
  Wide Web}, WWW ’07, page 697–706, New York, NY, USA, 2007. Association
  for Computing Machinery.

\bibitem{tansel1993temporal}
Abdullah~Uz Tansel, James Clifford, Shashi Gadia, Sushil Jajodia, Arie Segev,
  and Richard Snodgrass.
\newblock {\em Temporal databases: theory, design, and implementation}.
\newblock Benjamin-Cummings Publishing Co., Inc., 1993.

\bibitem{trivedi2017know}
Rakshit Trivedi, Hanjun Dai, Yichen Wang, and Le~Song.
\newblock Know-evolve: Deep temporal reasoning for dynamic knowledge graphs.
\newblock In {\em International Conference on Machine Learning}, pages
  3462--3471. PMLR, 2017.

\bibitem{upadhyay2018deep}
Utkarsh Upadhyay, Abir De, and Manuel Gomez-Rodriguez.
\newblock Deep reinforcement learning of marked temporal point processes.
\newblock {\em arXiv preprint arXiv:1805.09360}, 2018.

\bibitem{wijaya2015spousal}
Derry~Tanti Wijaya, Ndapandula Nakashole, and Tom Mitchell.
\newblock “a spousal relation begins with a deletion of engage and ends with
  an addition of divorce”: Learning state changing verbs from wikipedia
  revision history.
\newblock In {\em Proceedings of the 2015 conference on empirical methods in
  natural language processing}, pages 518--523, 2015.

\bibitem{zaveri2016quality}
Amrapali Zaveri, Anisa Rula, Andrea Maurino, Ricardo Pietrobon, Jens Lehmann,
  and Soeren Auer.
\newblock Quality assessment for linked data: A survey.
\newblock {\em Semantic Web}, 7(1):63--93, 2016.

\end{thebibliography}

\end{document}